\def\year{2022}\relax
\documentclass[letterpaper]{article} 
\usepackage{aaai22}  
\usepackage{times}  
\usepackage{helvet}  
\usepackage{courier}  
\usepackage[hyphens]{url}  
\usepackage{graphicx} 
\usepackage{amsmath}
\usepackage{amssymb}
\usepackage{amsthm}
\usepackage{array}

\usepackage{natbib}  
\usepackage{caption} 
\DeclareCaptionStyle{ruled}{labelfont=normalfont,labelsep=colon,strut=off} 
\usepackage{algorithm}
\usepackage{algorithmic}

\usepackage{newfloat}
\usepackage{listings}
\floatstyle{ruled}
\newfloat{listing}{tb}{lst}{}
\floatname{listing}{Listing}
\title{GLocal: Global Graph Reasoning and Local Structure Transfer\\for Person Image Generation}

\begin{document}

\maketitle

\begin{abstract}
In this paper, we focus on person image generation, namely, generating person image under various conditions, e.g., corrupted texture or different pose. To address texture occlusion and large pose misalignment in this task, previous works just use the corresponding region's style to infer the occluded area and rely on point-wise alignment to reorganize the context texture information, lacking the ability to globally correlate the region-wise style codes and preserve the local structure of the source. To tackle these problems, we present a GLocal framework to improve the occlusion-aware texture estimation by globally reasoning the style inter-correlations among different semantic regions, which can also be employed to recover the corrupted images in texture inpainting. For local structural information preservation, we further extract the local structure of the source image and regain it in the generated image via local structure transfer. We benchmark our method to fully characterize its performance on DeepFashion dataset and present extensive ablation studies that highlight the novelty of our method.
\end{abstract}

\begin{figure}[!t]
	\centering
	\includegraphics[width= 0.8\linewidth]{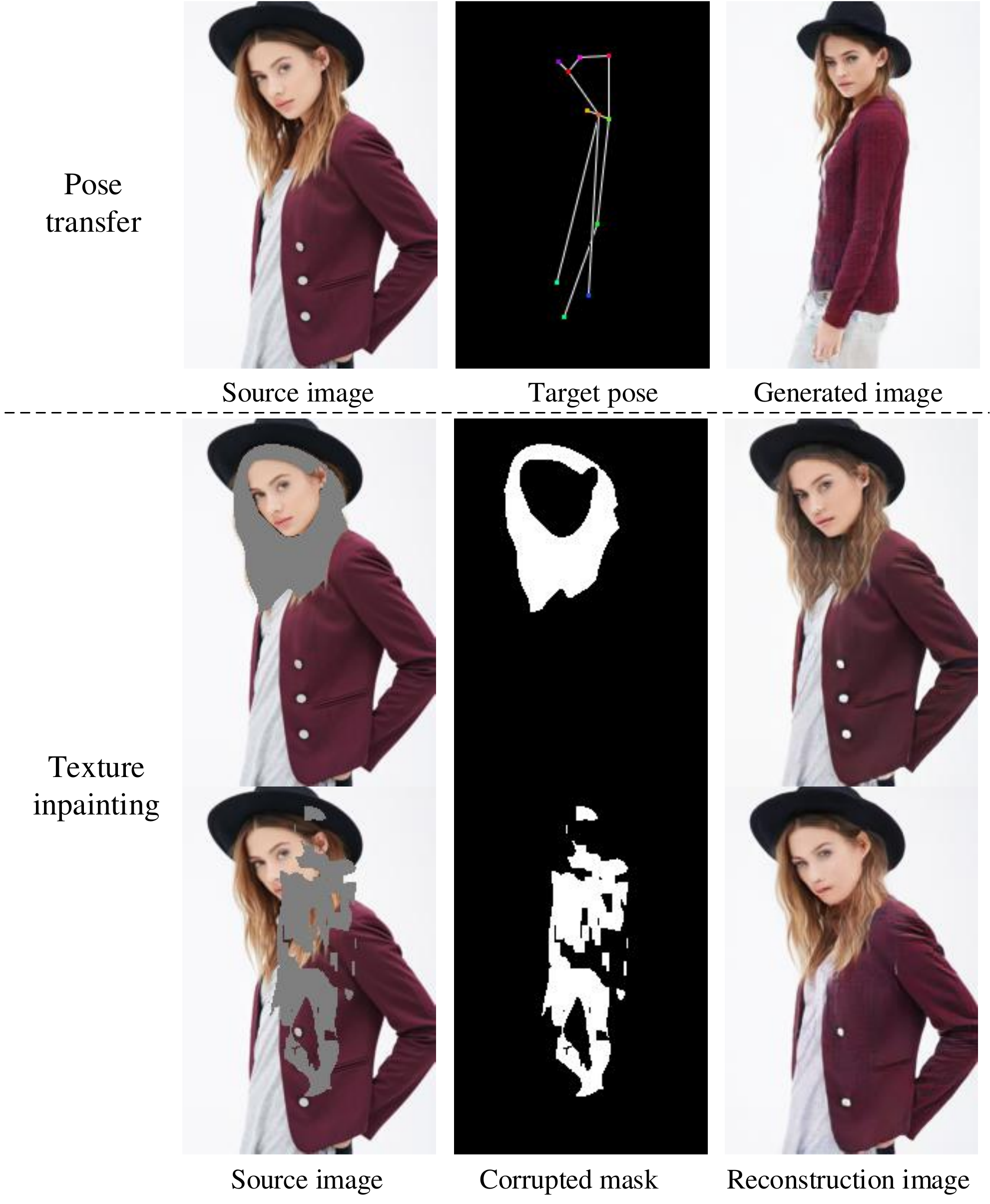}
	\caption{{The illustration of our GLocal, which can tackle tasks like pose transfer and texture inpainting.}}
	\label{fig:overview}
\end{figure}

\section{Introduction}
\noindent Person image generation aims to generate realistic person image with target pose while keeping the source appearance unchanged, which is a popular task in the community of computer vision with applications like image editing and image animation. Perennial research efforts have contributed to impressive performance gain on challenging benchmark. \\
\indent Essentially, the person image generation involves the deformation of human body in 3D space, which poses an ill-posed problem given just the 2D image and pose inputs. The limitation of 2D modeling nature leads to tough challenges due to the following difficulties: (i) the prediction of invisible parts in the source caused by occlusion or corruption as shown in Figure~\ref{fig:overview}, (ii) retaining the spatial structure of visible source texture. The exploration of these issues has contributed to the development and progress of this field.\\
\indent Existing works like~\cite{ma2017pose,pumarola2018unsupervised,chan2019everybody} propose solutions within the common image-to-image translation framework, which directly feed the conditioning pose and human image as encoder-decoder inputs. Thus it is unable to utilize the appearance correspondence between the input and target image. To achieve better rearrangement of source appearance, warping-based~\cite{dong2018soft, siarohin2018deformable,tang2021structure, li2019dense,liu2019liquid,ren2020deep} methods have been proposed. However, these operations struggle in recovering structural details by solely relying on the point-wise mapping and incur a defect when dealing with unmatched occlusion regions. Recent works~\cite{men2020controllable, zhang2021pise} apply semantic normalization techniques such as AdaIN~\cite{men2020controllable}, SEAN~\cite{zhu2020sean} to the semantic details restoration in the target. However, the non-discriminatory treatment of the occluded and non-occluded regions introduces misleading information and its capability to preserve structural details is not fully exploited.\\
\indent This paper presents \textbf{GLocal}, a scheme to preserve the semantic and structural information with \textbf{G}lobal Graph Reasoning (GGR) and \textbf{Local} Structure Transfer (LST), respectively. Inspired by the fact that the invisible area estimation can be facilitated by visible area (e.g. if the hand is unseen in the source image, its color can still be estimated by the neck skin), we manage to estimate the occluded region style by modeling the internal relationships among different semantic regions. The relationships are built with a graph architecture based on the linkage of each part, whose graph nodes are filled with per-region styles. Thereby the invisible area's style representation can be inferred by our GGR. After acquiring the global statistics of the source, we seek to reproduce source's local structure with LST since local structure can indicate how the image details are organized and is better than isolated points. We achieve this by predicting style parameters of the source from local correlation map and transferring the structural information of the source style to the generation features with LocConv. Our contributions can be summarized as follows:
\begin{itemize}
\item We propose GGR module to estimate the occluded region style with global reasoning in person image generation, which conveys visible style information to the invisible area along with the graph structure according to the connectivity of human body.
\item We design LST module to extract local structure and transfer it into the generated result, which sharpens the generation result and leads to better detail conservation.
\item Extensive experiments show that our method achieves superior performance on the challenging DeepFashion dataset both qualitatively and quantitatively.
\end{itemize}

\begin{figure*}[t]
	\centering
	\includegraphics[width=0.9\linewidth]{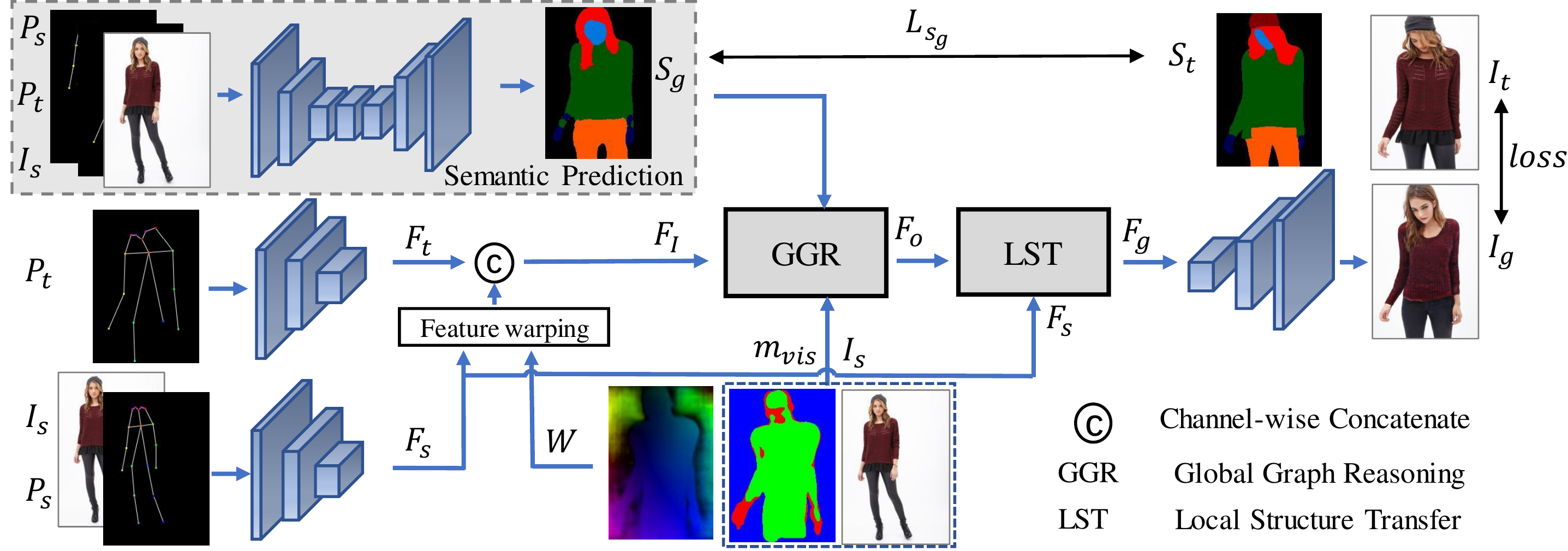}
	\caption{An overview of our GLocal framework. We first predict the target segmentation map $S_g$ from Semantic Prediction. Then we get the \textit{warping flow} $W$ and \textit{visibility map} $m_{vis}$ from a pretrained 3D flow estimation network~\cite{li2019dense}. We mark the visible and invisible positions with green and red in $m_{vis}$ respectively. After fusing the encoded target and warped source features, we can inject the source style into relevant regions within GGR module precisely. Finally, in the LST module, we transfer the local structure from source to generated result and send it into the decode to reconstruct the final image.}
	\label{fig:framework}
\end{figure*}

\section{Related Work}
\subsection{Person Image Generation}
Person image generation is a valuable branch of the mainstream image generation field, which focuses on the human-specific generation task. As a milestone, ~\cite{ma2017pose} first shows that conditional GAN (cGAN) can generate desired person images. Extending this idea, several following works seek to improve cGAN's performance with techniques such as unsupervised training~\cite{pumarola2018unsupervised}, cycle training~\cite{tang2020bipartite}, coarse-to-fine strategy~\cite{wei2020gac}, and progressive training~\cite{zhu2019progressive,liu2020pose}, etc. However, these methods ignore the vanilla conditional GAN's incapability to model the relationship on the large pose misalignment. To better achieve spatial rearrangement, ~\cite{siarohin2018deformable} decomposes the translation between different poses into local affine transformations. Approaches such as  ~\cite{li2019dense, Wei_Xu_Shen_Huang_2021, Tang_Yuan_Shao_Liu_Wang_Zhou_2021, liu2019liquid, ren2020deep} turn to estimate the appearance flow to align the source appearance with the target. Although warping the source can alleviate the aforementioned problem, they cannot deal with the occluded-region estimation for their false assumption that pixel-to-pixel warping transformation exists between source and target. Recently, emerging modulation-based methods~\cite{zhang2021pise, lv2021learning} adopt the two-stage pipeline, which first predict the segmentation map to extract region-wise image styles, and then use them to guide region synthesis with corresponding style injection. Unfortunately, this strategy cannot extract appropriate styles for occluded regions, and thus still fails to synthesize them. Different from previous methods, our global graph reasoning module can mitigate the occluded texture generation issues and achieve better structural preservation of the source in the target. 

\subsection{Semantic Image Generation}
Conditioned on a semantic segmentation map, semantic image generation is a special form of general image generation, whose solution is commonly based on style control. Style is mainly defined as the statistical property of the image feature which can be spatial varying~\cite{park2019semantic}, region adaptive~\cite{zhu2020sean}, or class specific~\cite{tan2021effi,tan2021diverse}. To transfer the style from one to another, they model the style as the modulation scale/shift parameters. Specifically, SPADE~\cite{park2019semantic} predicts spatial-varying affine parameters which modulates reference image features in different locations and obtains more dedicated style injection and high fidelity generation results. Later, SEAN~\cite{zhu2020sean} enforces the per-region style extraction and modulation. Our method stems from this scheme and injects the desired style based on whether it is occluded or not, which achieves more precise semantic manipulation.

\section{Method}
\subsection{Overview}
We propose a novel framework GLocal for person image generation which is divided into three stages: semantic prediction, global graph reasoning, and local structure transfer. As shown in Figure~\ref{fig:framework}, the semantic prediction network can supply semantic layout for per-region style extraction and injection in the global graph reasoning stage. The local structure transfer module further controls the generation process with local structure modeling and transferring.
\subsubsection{Preliminary}
To align the source feature $F_s$ with the target $F_t$ and supply visibility guidance, we adopt the Intr-Flow~\cite{li2019dense} to learn the appearance warping flow $W$ and visibility map $m_{vis}$ by matching the 3D human body model. The visibility map can indicate whether it is visible or not in the target for each source position.
\subsection{Semantic Prediction}
Directly generating a person image solely from target pose map involves the estimation of human semantic and body details, which is troublesome since they are highly related to each other. To simplify the overall generation process, we turn to generate the target semantic map first, which is easier to estimate and can provide better semantic assistance for further texture detail complement.\\
\indent Our semantic prediction network adopts the vanilla pix2pix~\cite{isola2017image} architecture, which takes the source image ${I}_s$, source pose map ${P}_s$, and target pose map ${P}_t$ as inputs. The semantic map consists of 8 semantic classes, among them some classes occupying a large area comprise the majority of the loss, which leads to inefficient learning for small parts (e.g., shoes). Considering this, we propose to utilize the focal loss~\cite{lin2017focal} to alleviate the class imbalance problem. Mathematically, the focal loss in class-balanced semantic prediction is formulated as:
\begin{equation}
\mathcal{L}_{\mathrm{S}_g}=-\left(1-p_{t}\right)^{\eta} \log \left(p_{t}\right), p_{t}=
\left\{
\begin{aligned}
\mathrm{S}_g, & \text { when } \mathrm{S}_t=1 \\
1-\mathrm{S}_g, & \text { when } \mathrm{S}_t=0
\end{aligned}
\right.
\end{equation}
where $\mathrm{S}_t \in\{1,0\} $ specifies the ground-truth target segmentation map and $\mathrm{S}_g \in [0,1]$ denotes the predicted probability for the class with label $\mathrm{S}_t=1$.  $\eta$ acts as the tunable focusing parameter and thus the scaling factor $\left(1-p_{t}\right)^{\eta}$ can effectively focus on the hard classes, e.g. shoes or hands.

\begin{figure*}[t]
	\centering
	\includegraphics[width=0.95\linewidth]{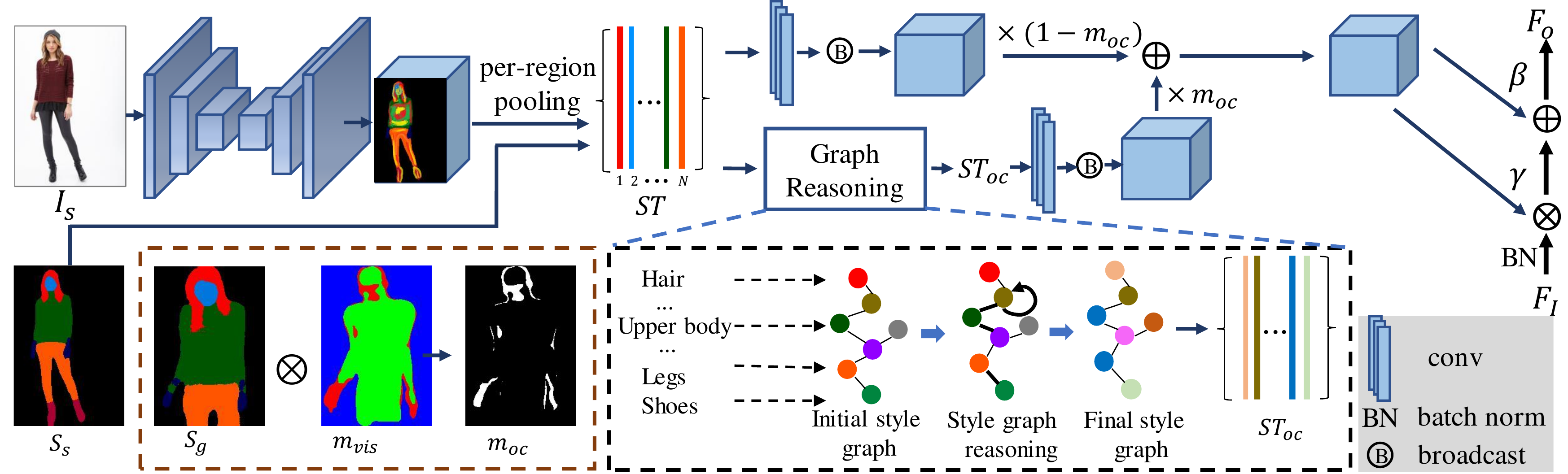}
	\caption{Architecture of our GGR (Global Graph Reasoning) module.}
	\label{fig:rgn}
\end{figure*}

\subsection{Global Graph Reasoning}
\label{sec:ggr}
As exemplified in Figure~\ref{fig:rgn}, to extract the style codes of the source for each region, we perform per-region average pooling on the encoded source features according to the source segmentation map $S_s$ and obtain the source's style code $ST \in R^{N \times 512}$, where $N$ denotes the number of semantic regions. To distinguish the occluded and non-occluded area for each region, we calculate the occlusion map $m_{oc}$ with predicted $S_g$ and visibility map $m_{vis}$ by element-wise product. For \textit{non-occluded} area, we inject each style into corresponding region. Then we perform per-style convolution and broadcast the convolved style codes to target regions. However, for \textit{occluded} area, directly generating with corresponding source region's style information may introduces irrationality. Thus we need to supply more accurate style representation for occluded area estimation with global style reasoning.
\subsubsection{Graph Modeling for Source Styles}
\begin{figure}[t]
	\begin{center}
		\setlength{\tabcolsep}{0.03cm}
		\renewcommand{\arraystretch}{0.5}
		\begin{tabular}{c}
			{\includegraphics[width=0.8\linewidth]{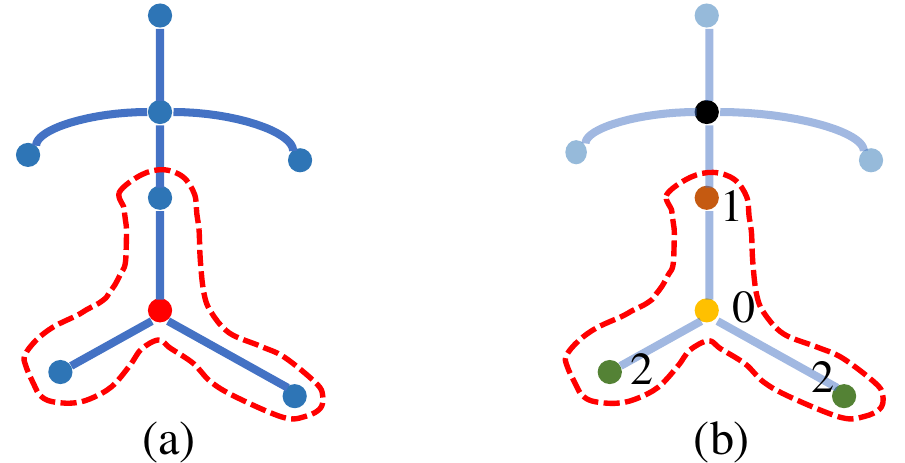}} 
		\end{tabular}
	\end{center}
	\caption{The illustration of nodes division strategy in our graph convolution. (a) The red dot represents the node where the convolution will take place and the nodes within the receptive field are surrounded by the a dotted line. (b) The neighbors are divided into three subsets 0, 1 and 2 according to its distance from the gravity center(black node).}
	\label{fig:graph}
\end{figure}

\indent  Since some body parts generally share similar appearance characteristics, e.g., the neck and the hand should be highly analogous in color. so it is natural to aid the occlusion estimation with global graph style propagation along with the human body structure. Given the encoded source feature style codes, we construct a relation graph with region-wise style code as graph nodes and natural connectivities in human body structure. Then the style feature nodes can be recurrently updated via graph propagation. Specifically, we construct a spatial graph $G=(V, E)$ on the N regions to feature the inter-region connection, where the node set $V=\left\{v_{i} \mid i=1, \ldots, N \right\}$ represents the all per-region style vectors. These nodes are connected with edges $E=\left\{v_{i} v_{j}\right\}$ according to the connectivity of human body structure. Then we perform spatial graph convolution to propagate the style information. The graph convolved node value at $\mathbf{x}$ can be written as:
\begin{equation}
{ST}_{oc}\left(v_{i}\right)=\sum_{v_{j} \in B\left(v_{i}\right)} \frac{1}{Z_i\left(v_ j\right)} {ST}\left(\mathbf{p}\left(v_i, v_ j\right)\right) \cdot \mathbf{w}\left(v_i, v_j\right)
\end{equation}
where the sampling function $\mathbf{p}: B\left(v_i\right) \rightarrow V$ is defined on the neighbor set $B\left(v_ i\right)= \left\{v_j \mid d\left(v_j, v_i\right) \leq D\right\}$ of node $v_i$ and $D$ is set to 1. The weighting function $\mathbf{w}\left(v_i, v_j\right)$ in graph convolution allocates a specific weighting value to each sampled node according to the subset it belongs, unlike the 2D convolution convolves its pixels according to the fixed square order. As illustrated in Figure~\ref{fig:graph}, we divide the neighbor set $B\left(v_i\right)$ of node $v_i$ into three subsets with centrifugal distance comparison. The mapping from nodes to its subset label is defined as:
\begin{equation}
r_{i}\left(v_j\right)=\left\{\begin{array}{ll}
0 & \text { if } e_{j}=e_{i} \\
1 & \text { if } e_{j}<e_{i} \\
2 & \text { if } e_{j}>e_{i}
\end{array}\right.
\end{equation}
where $e_i$ denotes the average distance from gravity center to node $i$. The convolution product of each subset is normalized by the balance term $Z_i\left(v_j\right)=\mid\left\{v_k \mid r_ i\left(v_k\right)=\right.\left.r_i\left(v_j\right)\right\} \mid$. \\
\indent After reasoning the occlusion area with style code graph, we can advance to the generation of conditioning feature map, whose occluded regions are now filled with globally reasoned source style features. By performing convolution and broadcasting on the conditional feature map, we obtain two sets of spatial-varying modulation parameters $\gamma$ and $\beta$. They act as the scale and bias to modulate the normalized feature map $F_{I}$ to get $F_{o}$.
\begin{equation}
	F_{o} = \gamma \times \textbf{BN}(F_{I}) + \beta
\end{equation}

\subsection{Local Structure Transfer}
\begin{figure}[t]
	\begin{center}
		\includegraphics[width=0.95\linewidth]{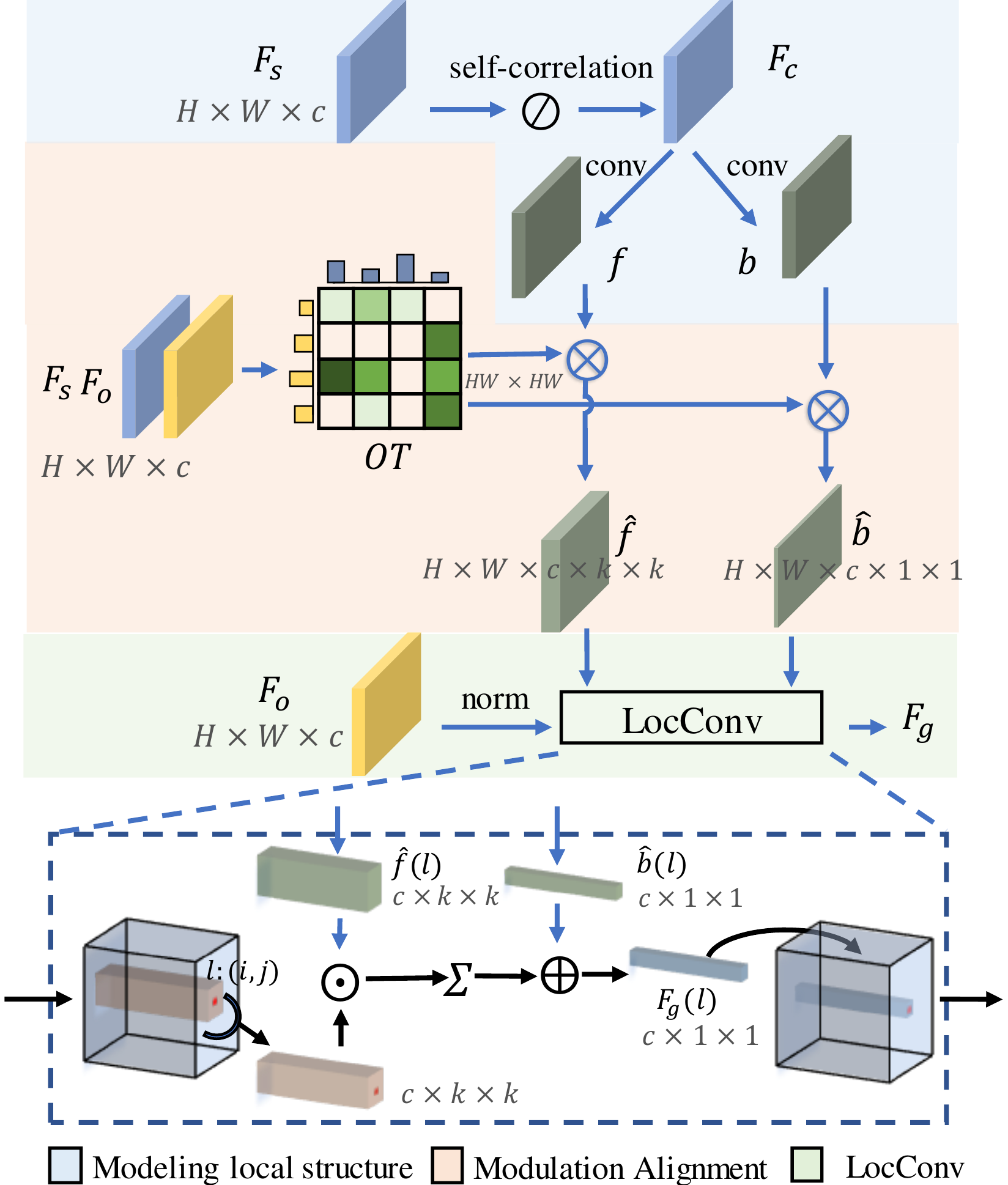}
	\end{center}
	\caption{Architecture of LST (Local Structure Transfer) module.}
	\label{fig:lsc}
\end{figure}

The GGR module can only capture the global statistics of the source and thus localized structure cannot be effectively preserved from source to target. As shown in Figure~\ref{fig:lsc}, to transfer the local structural context relationship, we first get local correlation map ${F}_{c}$, which models the local structure with self-correlation layer. Then we predict the full convolution kernels $f$ and bias $b$, since these kernels better distill local spatial structure. After aligning these parameters with the target by optimal transport, we perform LocConv to transfer the local structure of the source to the generation activations via convolution on the normalized target features.
\subsubsection{Modeling Local Structure}
Intuitively, the local structure of the feature map can be represented by the adjoining patch correlation patterns (i.e. the relationship of one patch with its neighbors). In light of this, we extract the local structural representation of source features with a self-correlation layer that perform multiplicative patch comparisons around each source position. Formally, given the source feature map ${F}_{s} \in \mathbb{R}^{c \times H \times W}$, our self-correlation layer calculates the correlation of two patches centered at $i$ and its neighbors $j \in \mathcal{N}(i)$ via vector product, which is defined as 
\begin{equation}
	\setlength{\abovedisplayskip}{5pt}
	\setlength{\belowdisplayskip}{5pt}
	\begin{aligned}
		c({i},j) &= \mkern-18mu \sum_{{p} \in [-r,r] \times [-r,r] } \mkern-36mu\langle{F}_s(i+{p}),{F}_s(j+{p})\rangle\\
		{F}_{c}(i) &= \mathop{Concat}\limits_{j \in \mathcal{N}(i) }({c(i,j)}), \vert\vert i - j \vert\vert \leq d
	\end{aligned}
\label{eq:patch_correlation}
\end{equation}
where the $Concat$ denotes the channel-wise concatenate. Note that for computational restriction, we just compute self-correlation $c(i,j)$  with neighbors whose distance from $i$ is less than $d$. After correlating the positions around $i$, we concatenate the correlations in channel to get the local structure representation ${F}_{c}\in \mathbb{R}^{(2d+1)^2 \times H \times W}$.
To better distill the local structure statistics and assist the structural transferring, we predict the spatial- and channel-varying modulation values including the 3D filter ${f} \in \mathbb{R}^{H \times W \times c \times (k \times k)}$ and bias ${b} \in \mathbb{R}^{H \times W \times c \times 1 }$ from the local structure representation ${F}_{c}$ via point-wise convolutions.
\subsubsection{Modulation Parameters Alignment}
To further align the modulation parameters with the target, we introduce the Unbalanced Optimal Transport (UOT)~\cite{zhan2021unbalanced} mechanism to match the encoded source $F_s$ and generation features $F_{o}$ with calculated transport plan ($TP$). To solve a optimal transport problem which aims to transform one collections of masses to another, we define the masses of source and generated features with dirac form: $\alpha=\sum_{i=1}^{n} \alpha_{i} \delta_{s_{i}}$ and $\beta=\sum_{i=1}^{n} \beta_{i} \delta_{{o}_{i}}$, where the $s_i$ and ${o}_i$ denote the positions of $\alpha_i$ and $\beta_j$ in the $F_s$ and $F_o$, respectively. Then the optimal transport problem can be formulated as:
\begin{equation}
	\begin{split}
		& OT(\alpha,\beta) = \mathop{\min}\limits_{TP} (\sum_{i,j=1}^{n}  {TP}_{ij}C_{ij}) = \mathop{\min}\limits_{TP} \langle C, TP \rangle \\
		&{\rm subject \ to} \quad (TP \vec{1}) = \alpha, \quad (TP^\top \vec{1}) = \beta \\ 
	\end{split}
	\label{formula_ot}
\end{equation}
where the cost matrix $C$ is formulated as $C_{i j}=1-\frac{s_{i}^{\top} \cdot {oc}_{j}}{\left\|s_{i}\right\|\left\|{oc}_{j}\right\|}$, giving the cost to move mass $\alpha_i$ to $\beta_j$. $TP $ denotes the transport plan of which each element $TP_{i j}$ denotes the quantity of masses transported between $\alpha_i$ to $\beta_j$. With the transport plan, we can warp $f$ and $b$ to get the aligned  modulation parameters $\hat{f}$ and $\hat{b}$, which is formulated as:
\begin{equation}
	\hat{f} = TP \cdot f, \hat{b} = TP \cdot b, TP  \in \mathbb{R}^{HW \times HW}
\end{equation}
\subsubsection{LocConv for Stucture Transfer}
  Unlike $1 \times 1$ kernel size adopted in the conventional point-wise modulation (e.g., SPADE~\cite{park2019semantic}), the $k \times k$ square filter of LocConv allows for modulating the generation acitvations with the local structure in the neighborhood  around point $l:(i,j)$ of feature map $F_o$. Then we can obtain generation feature map $F_g$ by modulating the aligned modulation filter $\hat{f}$ and bias $\hat{b}$ at $F_{o}$ in a spatial- and channel-varying way. The value of $F_g$ located at $l:(i,j)$ is defined as:
\begin{equation}
	\begin{aligned}
		F_g(l)&=\text {LocConv}(F_{o},l ; {\hat{f}}, \hat{b})  \\
		&= \sum_{{F_{o}}(p) \in \mathcal{N}({F_{o}}(l))} \hat{f}_{p}\left(\frac{{F_{o}}(p)-\mu_{{F_{o}}}}{\sigma_{{F_{o}}}}\right)+\hat{b}_p
	\end{aligned}
\end{equation}
where $\mu_{F_{o}}$ and $\sigma_{F_{o}}$ represent the channel-wise mean and standard deviation of the $F_{o}$. Finally we can get the $F_g$ by enumerating positions in $F_{o}$.
\subsection{Learning Objectives}\label{loss}
After the semantic prediction network is trained with our proposed $\mathcal{L}(\mathrm{S}_g)$. The whole networks is then trained end-to-end with source image ${I}_s$, source pose ${P}_s$, and target pose ${P}_t$ as inputs. We encourage the generated target-posed image ${I}_g$ to close in the groundtruth target image ${I}_t$ in image and perceptual level. Thus we introduce following learning objectives to guide the training process.

\noindent\textbf{Pixel-wise Loss.} To generate more sharp images, we use $\mathcal{L}_{L1}$ to measure the pixel-wise fiderlity between the generated image${I}_{g}$ and the groundtruth$I_{t}$ in image pixel space.
\begin{equation}
	\setlength{\abovedisplayskip}{4pt}
	\setlength{\belowdisplayskip}{4pt}
	\mathcal{L}_{L1} = ||{I}_{g}-I_{t}||_1,
\end{equation}
\noindent\textbf{Perceptual Loss.} Besides the pixel-wise loss, we also adopt the perceptual similarity measurement in VGG-19~\cite{simonyan2014very} feature space with perceptual loss.
\begin{equation}
	\setlength{\abovedisplayskip}{4pt}
	\setlength{\belowdisplayskip}{4pt}
	\mathcal{L}_{perc}=||\phi_k({I}_{g})-\phi_k(I_{t})||_2^2,
\end{equation}
{where $\phi_k$ represents the neuron response at $k_{th}$ layer extracted with a pretrained VGG-19 model.}\\
\noindent\textbf{Adversarial Loss.}
Due to the great potential of GAN, we introduce the adversarial loss to encourage the generator to generate photo-realistic images in adversarial manner. The training objective for the discriminator $D$ and generator $G$ is caculated with:
\begin{equation}
	\setlength{\abovedisplayskip}{4pt}
	\setlength{\belowdisplayskip}{4pt}
	\begin{aligned}
		\mathcal{L}_{adv}(G,D)=&\mathbb{E}_{I_{s},I_{t}}[\log(1-D(G(P_t,I_{s},{S}_{t} )|I_{s},P_t))]
		\\+&\mathbb{E}_{I_{s},I_{t}}[\log D(I_{t}|I_{s},P_t)].
	\end{aligned}
\end{equation}
\noindent\textbf{Overall loss function.} The final learning objective is incorporated with above-mentioned losses, which is defined as:
\begin{equation}
	\setlength{\abovedisplayskip}{4pt}
	\setlength{\belowdisplayskip}{4pt}
	\mathcal{L}=\alpha_{S_g}\mathcal{L}_{S_g}+\alpha_{L1}\mathcal{L}_{L1}+\alpha_{perc}\mathcal{L}_{perc} + \alpha_{adv}\mathcal{L}_{adv},
\end{equation}
where $\alpha_{S_g}$, $\alpha_{L1}$, $\alpha_{perc}$, $\alpha_{adv}$ are the trade-off weights.


\section{Experiments}\label{exp}
In this section, we first introduce implementation details including dataset and image quality evaluation metrics. Then we perform extensive experiments to illustrate the superiority over the prevalent methods and verify the effectiveness of our model.
\subsection{Implementation Details}
We select DeepFashion~\cite{liu2016deepfashion} to evaluate our method for its diversity among human identity and coverage of different clothes. The DeepFashion dataset contains 52712 high resolution images which show various clothing styles and poses. All images are resized to 256x176, which is adopted by many previous methods. Following the strategy of~\cite{ren2020deep}, we obtain image pairs and split them into 101966 for training and 8570 for testing without overlap. To measure generated results from different aspects, we choose Structural Index Similarity (SSIM) and Peak Signal to Noise Ratio (PSNR) for image-level similarity measurement which are sensitive to the image quality of imperceptibility. For perceptual evaluation, Inception Score (IS) and Fréchet Inception Distance (FID) are introduced to calculate the feature-level and distribution-based distance between the InceptionNet-encoded features.
\begin{figure*}[!t]
	\centering
	\includegraphics[width= 0.9\linewidth]{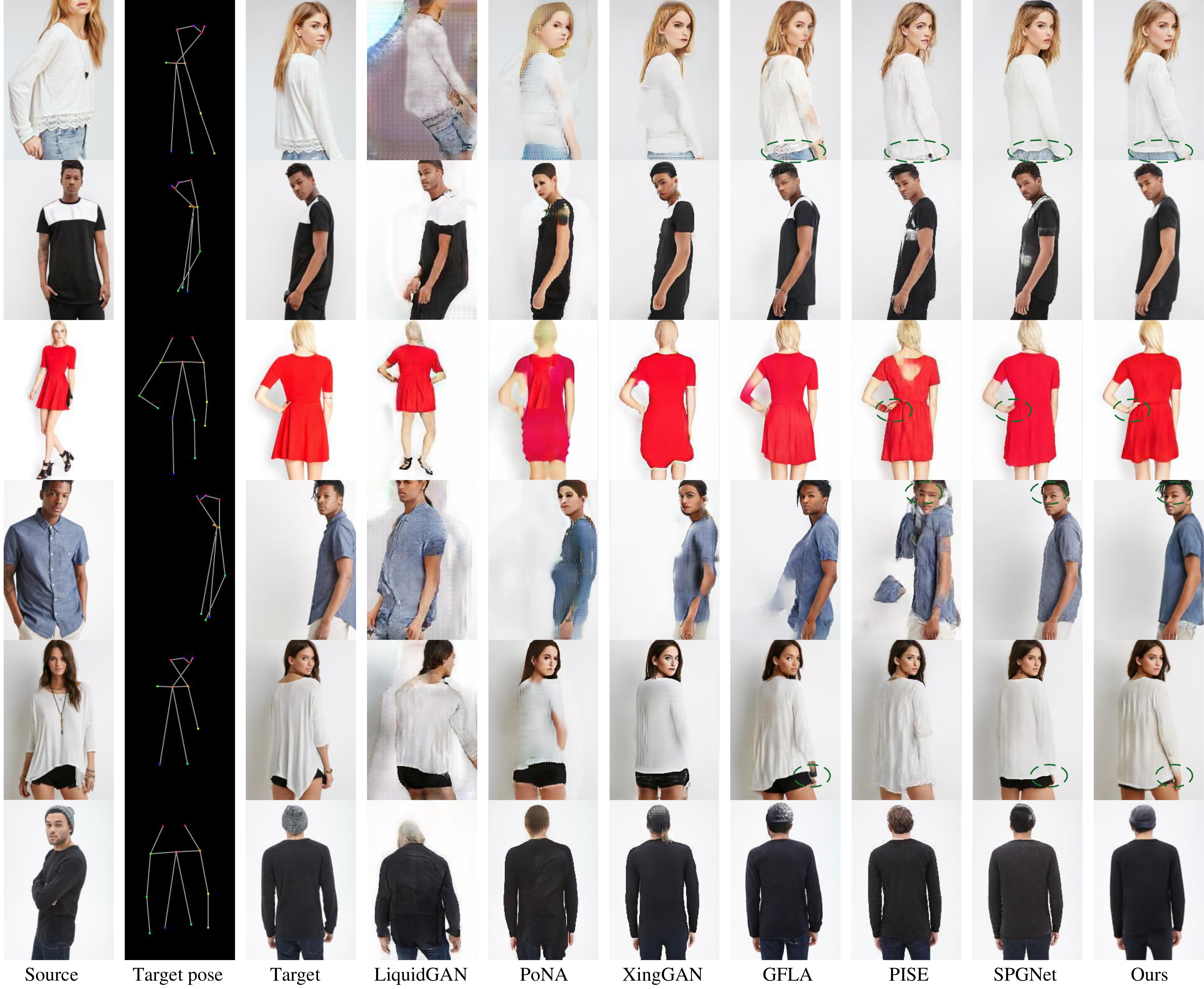}
	\caption{{Visual comparison with the competing methods on DeepFashion dataset. Best view it by zooming in the screen.}}
	\label{fig:comparison}
\end{figure*}
\subsection{Qualitative Analysis}
Figure~\ref{fig:comparison} presents the qualitative comparisons between several state-of-the-art methods and our model, which demonstrates the superiority of our method in clothes structure recovery, photo-realistic texture rendering, and distinct person identities preservation. The LiquidGAN~\cite{liu2019liquid} will generate weird human images when accurate 3D human modeling is not available. Methods such as PoNA~\cite{li2020pona}, XingGAN~\cite{tang2020xinggan}, and GFLA~\cite{ren2020deep} tend to generate unreasonable texture for lack of semantic guidance. Besides, the PISE~\cite{zhang2021pise} and SPGNet~\cite{lv2021learning} overlook the inter-region correlation and the distinction between occluded or non-occluded regions, resulting in visual artifacts like cluttered textures in the occlusion estimation. Different from all of them, our model adaptively predicts the invisible regions through graph-based region style reasoning. As circled in Figure~\ref{fig:comparison}, our model can maintain better shape consistency and generate more reasonable texture for invisible parts. Notably, benefited from our Local Structure Transfer module that can model and transfer local characteristics, our model can better preserve the hat structure as shown in the last row of Figure~\ref{fig:comparison}.
\begin{table}
	\setlength{\tabcolsep}{1.6mm}
	\centering
	\begin{tabular}{l|cccc}
		\hline \multicolumn{1}{c|} { Methods } & FID $\downarrow$ & IS $\uparrow$ & SSIM $\uparrow$  & PSNR $\uparrow$ \\
		\hline \hline LiquidGAN & $25.01$ & $\mathbf{3.56}$ & $0.613$ & $28.75$ \\
		PoNA & $23.23$ & $3.33$ & $0.774$ & $31.34$ \\
		XingGAN & $41.79$ & $3.23$ & $0.762$ & $31.08$ \\
		GFLA & $14.52$ & $3.29$ & $0.649$ & $31.28$ \\
		PISE & $\underline{13.61}$ & $3.41$ & $0.767$ & $\underline{31.38}$ \\
		SPGNet & $14.75$ & $2.99$ & $\underline{0.775}$ & $31.24$ \\
		GLocal(Ours) & $\mathbf{11.31}$ & $\underline{3.47}$ & $\mathbf{0.779}$ & $\mathbf{31.42}$ \\
		\hline
		\hline
		w/o FL & $15.88$ & $3.46$ & $\mathbf{0.779}$ & $\mathbf{31.42}$ \\
		w/o GGR & $17.39$ & $3.20$ & $0.769$ & $31.13$ \\
		w/o LST & $16.93$ & $3.43$ & $\underline{0.775}$ & $31.28$ \\
		\hline
		
	\end{tabular}
	\caption{Comparison with other state-of-the-art methods and variants on DeepFashion dataset. FID, IS, SSIM and PSNR are aforementioned metrics. $\uparrow$ and $\downarrow$ represent the higher the better and the lower the better. \textbf{Bold} and \underline{underlined} digits mean the best and the second best of each metric.}
	\label{tab:comp}
\end{table}

\begin{figure}[!t]
	\centering
	\includegraphics[width= 1.0\linewidth]{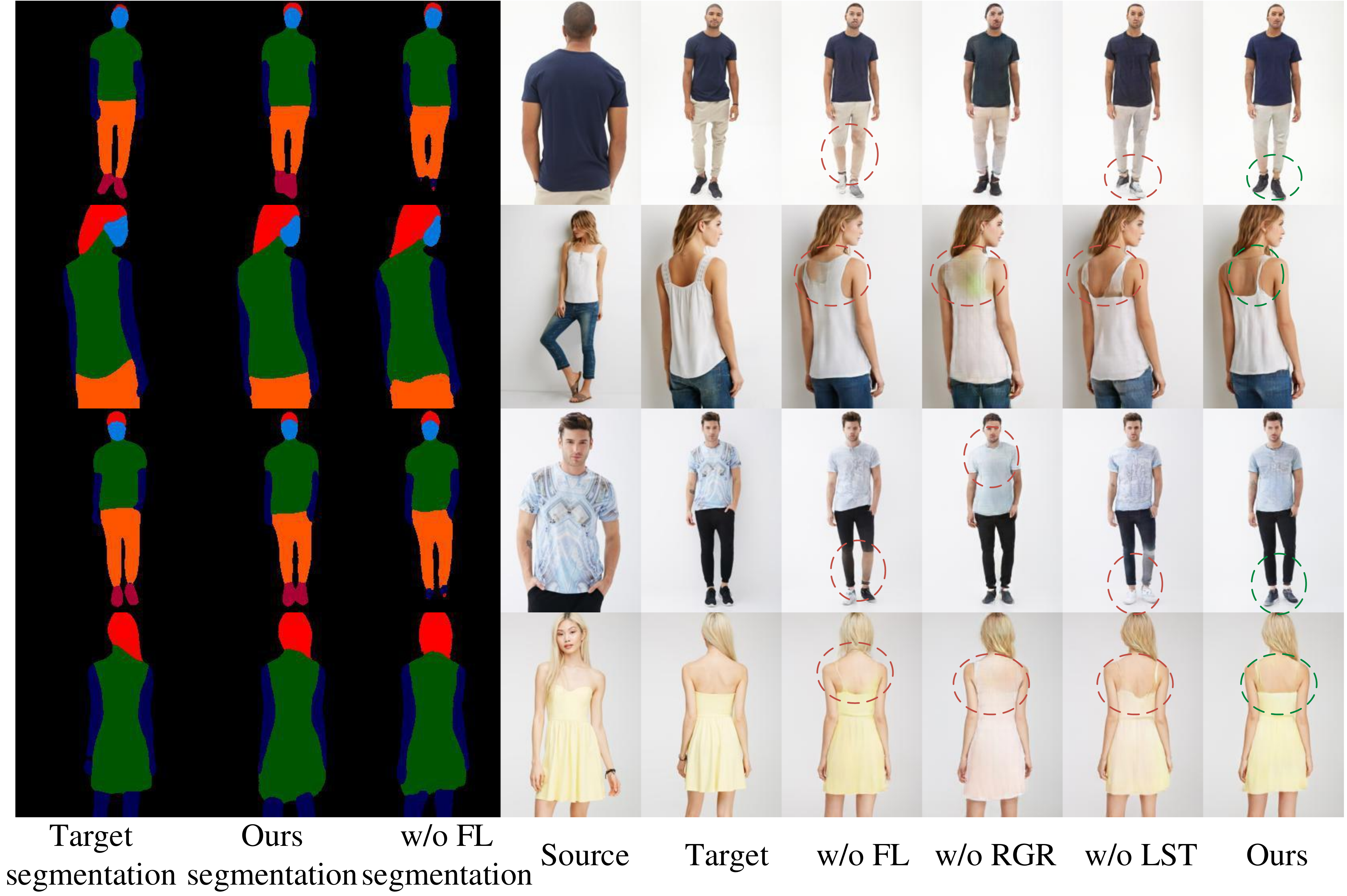}
	\caption{{The visual comparison of the variants and our full model. Best view enlarged on screen.}}
	\label{fig:ablation}
\end{figure}
\subsection{Quantitative Comparison}
We also compare state-of-the-art methods with our model by numerical cevaluation. Among all the models, our result reaches the highest SSIM and PSNR scores, which indicates that our model maintains the low-level statistical consistency with real images best. We further adopt deep metrics including FID and IS scores to measure the high-level perceptual consistency between our generated images and real images. As shown in Table~\ref{tab:comp}, our model leads the best in FID, which clearly shows the advantages to preserve the texture details. Besides, the IS score of our model surpasses most of prevalent models, which means that our GLocal can generate images with better quality and realism.

\subsection{Ablation Study}
We have trained several variant models to examine the effectiveness of our important components.

\noindent\textbf{w/o Focal Loss(w/o FL).} This variant predict the segmentation map with the vanilla multi-class cross-entropy loss.

\noindent\textbf{w/o Global Graph Reasoning(w/o GGR).} This variant removes the graph reasoning block and thus ignoring the distinction between occluded and non-occluded areas.

\noindent\textbf{w/o Local Structure Transfer(w/o LST).} This variant removes the local structure transfer mechanism and simply takes the features processed by global graph reasoning as the decoder input to generate the final image.

\noindent\textbf{GLocal (Ours).}This is the full model of our method.

From the quantitative result shown in Table~\ref{tab:comp}, we can verify the improvement of our three components. Compared with other variants, our full model outperforms them by a large margin in FID, which indicates that the cooperation of these components can generate more photo-realistic images. Besides, our full model also achieves the leading performance with the best IS, SSIM, and PSNR scores, which means that our full model can improve shape consistency, structure similarity, and pixel-level alignment with the real images. Intuitively, from the visual results in Figure~\ref{fig:ablation}, we can see that focal loss helps to achieve more dedicated semantic prediction since the less occupied region cannot be estimated under \textbf{w/o FL} setting (e.g. the segmentation map predicted by w/o FL lacks the sheet region.), which further leads to unrealistic texture rendering. For variants like \textbf{w/o GGR} and \textbf{w/o LST}, it fails to infer the occluded regions and preserve the structure information as depicted by the absurd texture performance, which is circled in Figure~\ref{fig:ablation}.

\begin{figure}[!t]
	\centering
	\includegraphics[width=1.0\linewidth]{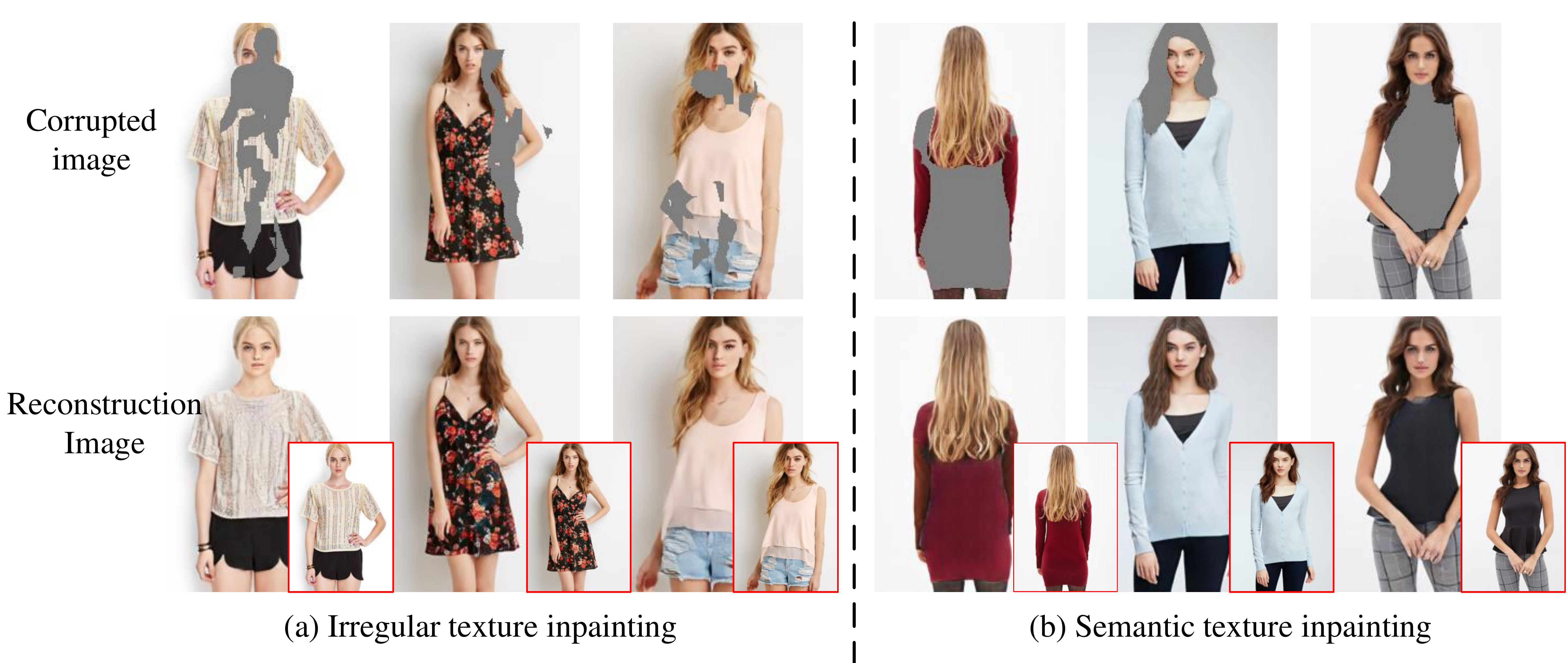}
	\caption{Texture inpainting application. We can perform image inpainting for missing regions by global graph reasoning in our GLocal model. The input are corrupted images which is masked with irregular or semantic masks and images in the red-box denote the ground truth images. }
	\label{fig:application}
\end{figure}
\section{Application}
Our GLocal model can also be extended to texture inpainting by inferring the invisible region in a corrupted source image. Given source image whose partial region is masked, the texture inpainting task can reconstruct the source image by filling in the missing region. The corrupted region in texture inpainting can be regarded as the occluded region in pose transfer since they both are invisible in the reference image. Based on this observation, we can utilize the region-wise graph reasoning to reconstruct the original image by calculating the style latent codes of the existing area and propagating the contextual style statistics into the missing regions with graph reasoning. The selection of mask shape divides our application into the following two categories.

\noindent\textbf{Irregular region inpainting.} The irregular region inpainting produces the inpainted result based on the irregular masked input, whose corrupted mask is provided by Irregular Mask Dataset\cite{liu2018image}. As shown in Figure~\ref{fig:application} (a), our method can generate correct structure and consistent textures for irregular regions with visible texture.

\noindent\textbf{Semantic region inpainting.} For semantic region inpainting, we randomly remove the texture for specific semantic regions (e.g.hair, pants). From the visual results in Figure~\ref{fig:application} (b), our approach is capable of recovering the content and rendering reasonable semantics. 

\section{Conclusion}
In this paper, we have presented a semantic-assisted person image generation framework to synthesis the target-posed source person or inpaint the corrupted image. Our approach models the semantic regions as a human skeleton-based graph and then infers the occluded target region's style with graph reasoning. To transfer the local structure from source to the generation features, the local property is learned with self-correlation and global-varying affine parameters are employed to modulate the generation activations. Extensive experiments and ablation studies have proven the superiority and effectiveness of our model.

\bibliography{aaai22}

\end{document}